\pdfoutput=1

\documentclass[11pt]{article}
\usepackage{algpseudocode}
\usepackage{algorithm}
\usepackage{commath}
\usepackage[final]{acl}
\usepackage{array, multirow}
\usepackage{nicematrix}
\usepackage{array,tabularx,booktabs}
\usepackage{times}
\usepackage{latexsym}
\usepackage{graphicx}
\usepackage{hyperref}
\usepackage[T1]{fontenc}
\usepackage[inline]{enumitem}
\usepackage[utf8]{inputenc}
\usepackage{amsfonts} 
\usepackage{amsmath,amssymb}
\usepackage{bbold}
\usepackage{nth} 
\usepackage{microtype}
\usepackage{sistyle}
\SIthousandsep{,}
\title{Interim Report on Human-Guided Adaptive Hyperparameter Optimization with Multi-Fidelity Sprints}

\author{Michael Kamfonas \\
  \texttt{mkamfonas@infokarta.com} \\
}

\newcommand{\model}{\mathcal{M}}
\newcommand{\fit}{\model_{F}}
\newcommand{\val}{\model_{V}}
\newcommand{\test}{\model_{T}}

\begin{document}
\maketitle
\begin{abstract}
This case study applies a phased Hyperparameter Optimization (HPO) process to compare multi-task natural language model variants that utilize multi-phase learning rate scheduling and optimizer parameter grouping. We employ short, intense Bayesian optimization sessions (sprints) that leverage multi-fidelity, hyperparameter (HP) space pruning, progressive halving, and a degree of human guidance, all within a constrained budget. Our work uses two Bayesian frameworks: the Optuna TPE sampler alongside the hyperband pruner and Scikit-Learn's Gaussian process minimization. Initially, we use efficient low-fidelity sprints to prune the HP space, i.e., eliminating less promising hyperparameter combinations. Subsequent sprints progressively increase their model fidelity and employ hyperband pruning for efficiency. Our framework allows the examination of optimal HP values at specific checkpoints, revealing hyperparameter trends through the training stages that inform scheduling decisions. Our study includes a prototype application that offers comparative visualizations and tools that facilitate hyperparameter space pruning. A distinctive feature of our approach is the integration of human judgment, enhancing the synergy between automated optimization and human intuition. 

A second aspect of our HPO approach is tuning threshold values to resolve classification probabilities during inference. Since these thresholds don't directly participate in training, we employ an iterative pre-validation meta-learner as a \emph{calibration} process.

Although we expect this framework to support a broad spectrum of cases, we demonstrate our methodology on JEREX-L, a collection of variants of the Joint Entity and Relation Extraction multitask model (JEREX by \citet{eberts-ulges-2021-end}), showcasing a reproducible, adaptable way to enhance machine learning model performance.

\end{abstract}

\pagenumbering{arabic}
\section{Introduction}

In natural language processing (NLP), optimizing model hyperparameters (HP) is a critical challenge, particularly for complex multi-task models where model options and variants must be compared. This study builds upon existing techniques and presents a phased Hyperparameter Optimization (HPO) framework that focuses the hyperparameter space on low-fidelity configurations before progressing to higher fidelities. We employ HPO cycles we call \emph{sprints}\footnote{Not to be confused with Agile sprints}. Each sprint entails a Bayesian optimization session targeting dimension refinement or optimal HP search. A human reviewer analyzes results and prunes the hyperparameter space as needed. An analysis and visualization tool with dimension pruning capabilities aids this process, for which we publish a prototype implementation. The methods we build upon are:

\paragraph{Multi-Fidelity Optimization:} We implement a granular strategy that adjusts the model's fidelity by subsetting data and varying training epochs. Although achieving lower fidelities by simply reducing training epochs is convenient, data subsetting acts as a cross-validation mechanism and results in more robust outcomes.

\paragraph{Hyperparameter Space Pruning:} By adjusting the hyperparameter space based on the outcomes of earlier low-fidelity sprints, we focus the search space for subsequent high-fidelity, more expensive sprints, leading to more efficient exploration.

\paragraph{Sprints and Threads:} Threads consist of progressively increasing fidelity sprints and focusing hyperparameter spaces ending with full-fidelity sprints with very few trials yielding the optimal models. All sprints in a thread must have the same model architecture and be initialized from the same checkpoint. Using a disciplined, phased approach in pruning hyperparameter space, applying multi-fidelity, and controlling top-N priming between sprints, we facilitate deep dives into specific configurations, ensuring thorough exploration and refinement.

We use two Bayesian optimization environments: Scikit-Optimize's gp\_minimize and Optuna's TPE sampler with the Hyperband pruner, and demonstrate the efficacy of our approach through a case study.

\section{Background and Related Work}
\subsection{JEREX-L Overview}

JEREX-L is a comparative study of enhancements to the original Joint Entity and Relation Extraction Model (JEREX) by \citet{eberts-ulges-2021-end}, an end-to-end extractor for entities and relations. The architecture consists of shared embedding and encoder layers, feeding four tasks, i.e., mention localization, coreference resolution, entity classification, and relation classification. We assume that the reader is familiar with the original paper. Only the subset of configuration options considered in this study is elaborated below.

    \paragraph{Encoder Model:} The original JEREX model uses \emph{bert-base-cased} \citep{Devlin2019BERTPO} as the encoder. 
    We generalized the framework to enable several auto-model transformers from Hugging Face \cite{wolf-etal-2020-transformers} through configurable arguments. Results in this article compare a second implementation that uses \emph{longformer-base-4096} \cite{Beltagy2020Longformer} by Allen AI, which allows longer input sequences and can accommodate all examples of the DocRED and re-DocRED datasets. 
    The Longformer architecture features a memory-efficient attention mechanism that can handle long passages at a nearly linear cost. It combines local (windowed) and global attention, which resonate well with spoken or written language, where semantic word linkages tend to diminish with distance, but occasional important exceptions exist and must be handled. 

    In our experiment labels, we use abbreviations \emph{Bert} and \emph{Long} to distinguish experiments performed with the two models.

    \paragraph{Optimizer Parameter Groupings:} 
    We compare two optimizer parameter partitioning schemes, one adhering to global rates, per the original article, and one with separate rates for different module groups.  \autoref{tab:paramGroups} describes these two grouping patterns.
\begin{table}
\centering
\begin{tabular}{lllr}
\textbf{GLOBAL} & \textbf{LR0} &  \textbf{L2} \\
\cmidrule(r){1-3}
\multirow[c]{11}{*}{All}  &\multirow[c]{2}{*}{Shared } &  No \\
 &  & Yes \\

\cmidrule(r){2-3} 
& \multirow[c]{2}{*}{Mention} & No \\
&                             &  Yes \\

\cmidrule(r){2-3}
& \multirow[c]{2}{*}{Coreference}  & No \\
& & Yes \\

\cmidrule(r){2-3}
 &  \multirow[c]{2}{*}{Entity} & No \\
&  & Yes \\

\cmidrule(r){2-3}
& \multirow[c]{2}{*}{Relation}   & No \\
& &  Yes \\
\cmidrule(r){1-3}
\end{tabular}
\caption{Taxonomy of model parameter blocks into Global or by task module (LR0). The last column captures the subdivision into whether modules are subject to L2 regularization.}
\label{tab:paramGroups}
\end{table}

\begin{description}
    
    \item[GLOBAL:] This is the scheme used in the original JEREX. One learning rate applies to all groups, and one weight decay rate applies to groups subject to L2 regularization\footnote{The choice of parameters excluded from L2 regularization is the same as in the 
    original paper, i.e., embeddings, normalization layers, and all biases. }. 
    \item[LR0-L2:] A scheme of five learning rates for five module groupings (LR0) and five weight decay rates for the respective modules subject to L2 regularization (when $L2=$Yes.) 
\end{description}

    \paragraph{Linguistic Augmentation:} The main component of the original representation of mentions and relations consists of contextual embeddings generated by the encoder transformer. We hypothesize that adding linguistic features like part-of-speech (POS) may enhance performance. We employ\emph{spaCy} \citep{honnibal-johnson-2015-improved,spacy2} to do this. Encodings of these linguistic features are applied a little differently for the two transformers we compare. For Bert, we extend the hidden representation output by appending POS encoding. For Longformer, we use nouns and proper nouns as global features and leverage the global attention mechanism. 

    We compare Bert with and without the POS option, while we always use POS-based global attention for Longformer models.

    \paragraph{Data Imbalance} The original model controls the proliferation of negative examples by limiting their sampling to a fixed count and favoring the more complex negative instances whose spans overlap with positive ones. We complement this approach with the option for \emph{Adaptive Asymmetric Loss (ASL)}, a function of prediction difficulty \cite{benbaruch2020asymmetric}. This method is a generalization of \emph{Focal Loss} \citep{Lin2020FocalLF} with asymmetry introduced by parameterizing positive examples differently than negative ones. Supplemental information about this method is included in the appendix.

    We compare configurations with and without the ASL option.

    \paragraph{Multitask Loss:} In JEREX, multitask loss is computed as the weighted sum of the four task losses. The original model used fixed weights as hyperparameters. We added an option for \emph{Uncertainty-based regularization}  \citep{Kendall2018MultitaskLU,meshgi-etal-2022-uncertainty}, whereby task loss weighting is dynamically adjusted based on the uncertainty reflected in the contributing task predictions. 

    We compare model configurations with the dynamic task loss option (DTL) versus fixed weighting.

    In addition to the configuration options we compare in the study, we made some changes that apply to all configurations. These include some code refactoring and equivalence transformations, facilitating upgrades of the Torch and Lightning frameworks, and performance and convenience enhancements. 
    
    We also made some changes that are not equivalent to the original model, that are listed below: 
    
    \paragraph{The Dataset:}
    The original model was trained and tested on DocRED, \cite{Yao2019DocREDAL}, a dataset designed for general document-level entity and relation extraction. What is challenging about DocRED is that it contains multi-sentence examples annotated for six named entity types and 96 relation classes, some of which require reasoning through multiple sentences to predict correctly. This makes multi-sentence relation extraction much more challenging than similar single-sentence datasets like TacRED \cite{zhang2017tacred}. We trained and tested JEREX-L on Re-DocRED \cite{tan2022revisiting}, an improved version of DocRED that fixed many incorrect or missing annotations in the original version. 

    \paragraph{Relation Thresholds:}
    We replace the original global threshold value used for all relation classes with a vector of individual, independently optimized values, one for each relation class.

    \paragraph{Representation Changes:}
    In the original JEREX model, edit distances in coreference resolution and token or sentence distances in relation classification were represented through embeddings, reducing them to discrete categorical values with no notion of order or continuity.  

        In JEREX-L, we adopt log-linear encoding, which offers a superior alternative to embeddings as it preserves continuity, order, and a more appropriate utility-scale. The essence of these distances lies in the nonlinear perception of affinity they signify between the endpoints, which is more suitably conveyed by the distance's logarithm. 


\subsection{Prior Work on Bayesian HPO}
Bayesian optimization with Gaussian Processes, a black box approach to global optimization problems, was formulated in the 70s and 80s \citep{Mockus1989BayesianAT} and plays a dominant role in the constellation of methods currently used for Hyperparameter optimization (HPO) of deep learning networks  \citep{bischl2021hyperparameter,snoek2012practical}. 

\paragraph{The Gaussian process} as defined by 
\citet{rasmussen:williams:2006} is a collection of random variables, any finite number of which have a joint Gaussian distribution. A GP is defined in terms of a mean and a covariance function. 
\begin{align*}
f(x) &\sim \mathcal{GP}(m(x), k(x, x')) &\\ 
\intertext{where:}
m(x) &= \mathbb{E}[f(x)] &\\
k(x, x') &= \mathbb{E}[(f(x) - m(x))(f(x') - m(x'))] &
\end{align*}

In an HPO setting, $f(x)$ represents the objective function evaluated at point $x$ of the hyperparameter space. 

There are various choices for the covariance function implementation. The  \emph{gp\_minimize} function from the \emph{Scikit-Optimize} package defaults to the widely used Mat\'{e}rn kernel \citep{Matern1960SpatialV,rasmussen:williams:2006}.

During the Tuning process, iterations on the objective function $f(x)$ continuously improve the Gaussian process, enhancing point selection for subsequent trials. The selection is made using an acquisition function, which can exploit the accumulated knowledge to make better choices or explore new uncharted areas by randomly selecting points. Acquisition functions \citep{frazier2018tutorial} are designed to support different strategies, such as Probability Improvement (PI), which seeks a higher probability of beating the best score; Expected Improvement (EI), which maximizes the magnitude of improvement; and Lower Confidence Bound (LCB), which minimizes regret\footnote{Regret is the difference between the best possible outcome and the outcomes obtained by the algorithm over time. Its minimization is achieved by balancing exploration-exploitation.} throughout the optimization. We use the \emph{gp\_minimize} implementation, which defaults to a probabilistic algorithm that selects one of the three acquisition functions at each iteration.

The Gaussian process is cubic in the number of data points and does not scale well when too many iterations are involved, particularly as the number of parameters increases. The \emph{Tree-structured Parzen Estimator (TPE)} method \cite{Bergstra2011AlgorithmsFH, Bergstra2012RandomSF, snoek2012practical,Watanabe2023TreestructuredPE} is a more efficient alternative. Unlike conventional random search, TPE iteratively updates two distributions, $l(x)$ and $g(x)$, based on the outcomes of previous trials. Using a quantile-based threshold, it first categorizes trials into "good" and "bad" based on their objective function values. The $l(x)$ distribution is then updated to model the distribution of hyperparameters from the good trials, while $g(x)$ models the distribution from the bad trials using Parzen window estimators or kernel density estimators. This process ensures that regions of the hyperparameter space associated with better performance have a higher density in $l(x)$, influencing the algorithm to sample new points from these promising regions. The decision to sample new hyperparameters is based on maximizing the ratio $\frac{l(x)}{g(x)}$, which directs the search towards areas where the likelihood of improvement is greater, leveraging the knowledge gained from the outcomes of prior trials. A popular framework supporting TPE is Optuna \cite{akiba2019optuna}, which we use in our implementation.

\paragraph{Progressive Halving} is another cornerstone of modern HPO frameworks \cite{li2018hyperband}. It originates from the insight that not all hyperparameter configurations need full resource allocation to be evaluated effectively, and its theoretical underpinnings build upon the one-arm-bandid problem  \citep{jamieson2016non}. It is a budget-saving strategy that initially allocates minimal resources to many candidates, progressively increasing the investment in those showing promise at the expense of those who don't. This approach aligns with the ``racing'' algorithms principle, where underperforming candidates are quickly eliminated, allowing for more efficient exploration of the hyperparameter space. Progressive Halving stands out for its ability to handle large hyperparameter sets with a limited computational budget.


\paragraph{Multi-fidelity Techniques} leverage the concept that evaluations of the objective function can be performed at different levels of fidelity, offering a trade-off between computational cost and accuracy. These techniques apply to both surrogate-based optimization (such as GP) and bandit-based strategies (such as TPE,) using cheaper, low-fidelity evaluations to guide the search in the hyperparameter space \cite{kandasamy2017multi}. Theoretical foundations suggest that a well-calibrated balance between low and high-fidelity evaluations can significantly accelerate the optimization process by prioritizing computational resources towards the most promising regions of the hyperparameter space \cite{poloczek2017multi}. The prevailing method for achieving multi-fidelity in HPO is varying the number of epochs. Subsetting the training/validation dataset is another way, but it lacks the natural and seamless transition from low to high fidelity that the epochs allow.


\paragraph{Hyperband} proposed by \citet{li2018hyperband} enhances the Progressive Halving algorithm with a systematic approach to dynamically allocate and adjust computational resources among a set of hyperparameter configurations. Hyperband optimizes the exploration-exploitation trade-off by employing a bandit-based strategy that evaluates, prunes, and focuses on promising configurations across different scales of resources. This method elegantly combines the idea of early stopping (pruning) with an adaptive resource allocation mechanism, optimizing the efficiency of the HPO process, especially in the presence of larger hyperparameter spaces and computational budget constraints.


\paragraph{Multi-phase Learning Rate Schedulers,} in conjunction with the optimizer, adjust the learning rate (and other regularization parameters) according to a predefined schedule over the training process. These schedulers adjust HPs through warmup-and-decay cycles over the length of the training. However, when integrated with HPO, two problems may arise. (a) Restart state distortion is a side-effect of the complexities of restarting a suspended training session from a checkpoint, whereby the optimizer, scheduler, and possibly other components are not properly reset to the correct state. (b) Cycle compression distortion arises when schedules meant to unravel over long multi-epoch training cycles are compressed into a short interval, e.g., one epoch, thus drastically accelerating the effective rate changes of HP values. These distortions may not fully account for the dynamic interplay between learning rate scheduling and model performance across all phases of the training regimen \cite{loshchilov2016sgdr}. Addressing this challenge involves developing HPO strategies that ensure the optimization process remains aligned with its true full-fidelity trajectory.

The current state-of-the-art methods tend to implement multi-fidelity through epoch reduction, thus increasing the fidelity of the most promising configurations. However, low fidelity achieved by subsetting data provides more leverage and we use it for pruning dimensions. A Two-Step HPO method proposed by \cite{Yu2023TwostepHO} demonstrates that starting with small training and validation data subsets and progressively using more data is an effective and economical way to narrow choices and value ranges and converge, if not to a global,  at least to a decent local minimum. However, the authors caution against using the proposed approach with adaptive tuning techniques, such as Bayesian HPO, because selections depend on history and may be based on different fidelity trials. We heed this advice by introducing sprints, i.e., autonomous sessions with constant fidelity, passing historical experience through pruned dimension spaces, and re-evaluating data points when crossing fidelity sprint boundaries.  

\subsection{Related Work on Thresholds}

Appropriate mapping from classifier output to scores is crucial for model selection and effective inference in deployment \cite{HernndezOrallo2012AUV}. Thresholds are the most common mechanisms for facilitating this mapping. 
Prediction performance in multi-label classification is often highly sensitive to the choice of thresholds used to partition the spectrum of classification probabilities into binary True or False decisions for each label.  Each instance can be associated with multiple labels, and the challenge lies in accurately predicting the subset of relevant labels for each instance. Binary classification is a trivial case of the same concept with only one label involved.

Two of the many approaches proposed for improving prediction in multi-label classification \citep{Yang2001ASO} are:

\paragraph{RCut:} This method involves ordering the labels by their predicted probabilities and using a parameter ($k$) to select the top-$k$ labels as the predicted label set for a given instance.
\paragraph{SCut:} This approach ranks instances based on each label's relevance probability. Iterating over these probabilities, the algorithm seeks the cutoff value that best separates positive from negative outcomes for each label, aiming to maximize the performance metric of interest.

\paragraph{}The SCut method can be applied to training or validation datasets or even to test data as ``Oracle'' for comparing algorithms \citep{Fallah2022AdaptingTF}. Our calibration methodology uses validation data exclusively, and the resulting thresholds are used in testing or inference. 
We observed that when applied early in the fine-tuning process, where positive labels are grossly underrepresented, the binary SCut algorithm can set the optimal threshold to such a low value that almost no candidate instances get filtered out, thus significantly slowing down the evaluation. We consequently take note of a solution inspired by the SCutFBR.1 variant of SCut \cite{Yang2001ASO}. According to a comparative study by \cite{Fan2007ASO}, the SCutFBR.1  handles under-represented labels better by setting the threshold to the highest probability for the label if the micro-$F_\beta$ score falls below some low limit. Our algorithm, instead of basing this on the micro-$F_\beta$ score,  prevents early threshold estimations from falling lower than a chosen low bound.

While threshold-based methods are the bedrock of multi-label classification, they inherently assume that the relevance of labels can be mapped to a binary decision based on a static cutoff point. This assumption does not account for any potential variability in the number of applicable labels across instances, often leading to suboptimal precision and recall balance. In response to this limitation, \cite{Fallah2022AdaptingTF} proposed the \emph{N Highest Activations (NHA)} method, shifting the paradigm from static thresholds to dynamic, instance-specific label prediction. Instead of deriving label probabilities, the NHA method adds a parameter $k$ to the classification layer that captures the optimal number of labels for each instance, which is optimized according to the mean absolute error (MAE) loss. It then uses the RCut method to extract the top-$k$ labels. This approach addresses the challenges of over- and under-tagging inherent in threshold-based methods. It offers a more nuanced pathway to achieving higher accuracy in multi-label classification scenarios with uneven or fluctuating label participation. While this approach would be worth exploring, currently, JEREX and JEREX-L employ threshold-based (SCut) techniques.



\section{Methods}
 Our implementation leverages the \href{https://scikit-optimize.github.io/stable/auto_examples/bayesian-optimization.html}{skopt} optimization core framework for GP-based optimization and the \href{https://optuna.org/}{Optuna} framework \cite{akiba2019optuna} for TPE selection with Hyperband pruning.

We use the following three-phase method as our baseline HPO process:

\paragraph{Phase 1: Defining the HP Subpace}
The primary objective of Phase 1 is to efficiently delineate the hyperparameter (HP) space with minimal computational expenditure. Utilizing \textit{gp\_minimize} from Scikit-Optimize, we conduct 90 to 120 one-epoch iterations on a low-fidelity model,  with half of these iterations employing random search to ensure broad coverage. Since the training cycle is highly compressed, the learning rate scheduler is intentionally disabled to prevent distortion of the performance landscape. Consequently, the HPs reflect the optimal effective values the scheduler should produce had it been active. The outcome we seek from this phase is a refined HP subspace that envelopes the global or a decent local optimum. Human-guided adjustments, aided by a visualization and space-pruning tool, can be made to the ranges of each dimension based on the performance of the top 10 iterations, plus a margin. Optionally, multiple such sprints may increase the probability that more local minima are uncovered. Our experiments were based on one such sprint.

\paragraph{Phase 2: Warmup Tuning in the HP Subspace}
Phase 2 delves into hyperparameter tuning within the previously established HP subspace,  focused on optimizing the learning and weight decay rates that maximize performance at the end of the warming period. The key objective is to determine optimal hyperparameter sets for the cyclic lr-scheduler, ensuring that the effective values it produces maximize scores at the end of the warmup period. We conduct between 60 and 90 iterations using \textit{gp\_minimize} for $\lceil w N_\text{max} \rceil$ epochs where $w$ and $N_\text{max}$ are the parameters \text{pct\_start} and \text{max\_epochs} of the scheduler. The \emph{max\_epochs=25} mirroring full-fidelity conditions, to prevent distortion of the performance landscape. An early stopping mechanism triggers trial termination at the end of the warming period. This phase yields optimal hyperparameter configurations, including the scheduler's crucial maximum learning rate (\emph{max\_lr}) and warming period (\emph{pct\_start}), as well as the optimizer's weight decay (\emph{weight\_decay}). 

\paragraph{Phase 3: Full-Cycle HPO }
The culmination of our HPO process, Phase 3, aims to integrate the full training regimen, including training, validation, and post-training threshold calibration, employing the full-fidelity model. Leveraging Optuna's TPE and Hyperband pruning mechanisms, the study is initially primed with the best 3 to 5 HP configurations derived from Phase 2, followed by additional trial generation. This phase may optionally accept restricted HP space definitions, from prior phases,  which can be manually adjusted upon evidence of superior ranges. The result of this step is a short list of optimal configurations and trained model checkpoints based on validation data. 

\paragraph{}Each of these phases is implemented through ``sprints'' corresponding to single gp\_minimize or Optuna study sessions. A sprint can be extended by adding more iterations or trials as long as the model configuration, the set of optimized hyperparameters, its fidelity, and the hyperparameter space are the same. The sprint outcome includes the resulting HP configurations and, through human review and guidance, a potentially adjusted subspace, which may be pruned to more restrictive boundaries for each dimension than the initial space it started with. In certain cases, the adjustments may broaden the ranges when optimal parameter values flock around one of the boundaries, indicating the possibility of optimal values being outside the defined range. Another extreme way of pruning is freezing certain HPs. This is useful in cases where a value is confidently determined, or certain conditions make the parameter irrelevant. An example of the latter case is when we disable scheduling for phase 1, the warmup period becomes irrelevant, and its presence in the set of optimized parameters only confuses the optimization process. 

In the discussion that follows, we refer to three types of ``configurations'':
\begin{description}
    \item[Model Configuration] encompasses the architectural blueprint of the model, its initialization regime, parameter grouping scheme, and computational strategies employed, such as key metrics, loss functions, or parameter-efficient methods. 
    \item[Training Configuration] pertains to the dynamic aspects of how the model is trained through the HPO iterations (trials), the hyperparameters being optimized and their values 
    \item[Optimization Configuration] in our study's context pertains to the dimensional space definition of hyperparameters subject to optimization and the model's fidelity. Fidelity is usually controlled by the number of epochs or data subsetting but can also involve changes in the model configuration, such as disabling lr-scheduling. The history of trials or best-performing subsets of this history used in the selection of new trial points is also part of the optimization configuration.
\end{description}

We now clarify what we mean by ``Initialization Compatibility:'' The objective function used during the HPO process initializes the model to the same consistent state at the beginning of each iteration. This state typically reflects the initial pre-trained weights of the encoder (embedding and transformer layers) and randomly initialized task weights. It is possible, however, to load the model from a later checkpoint, such as at the end of the warmup period or a mid or late point in the training trajectory. Performing single-epoch HPO sprints with the scheduler disabled allows us to discover what the optimal effective learning and weight decay rates, or other HPs, are optimal for that phase of the training cycle, and can be used to inform schedule finetuning.  Generally speaking, optimal effective hyperparameter values\footnote{Such as the learning rate in effect set by the scheduler}  may change dramatically over the trajectory of training, so the epoch and global step of the checkpoint used for model initialization is included in the label identifying a sprint. Sharing HP subspaces or trial sets across different initialization points is strongly discouraged.

We use specific guidelines for how trial history and adjusted HP spaces can be shared across sprint boundaries:

\begin{description}
    \item[Sharing Pruned Spaces:] A sprint may use an adjusted HP space based on trial history of a previous sprint with the same model configuration. This is a way to pass HP space definitions delineated by lower fidelity sprints to more costly ones without sharing trial data. Although Bayesian methods exploit historical knowledge to suggest new trial HP sets anyway, focusing the envelope of dimension ranges makes exploration more efficient. The risk of constraining ranges too much, and missing a potential optimal minimum in an area not yet explored is a legitimate concern when considering how aggressive dimension pruning should be.  
    \item [Warm Priming:] This is when top-N HP value sets and the respective scores from previous sprints are added to new sprints. In these situations, the new sprint trusts the prior score and includes the trial in its history, exploiting it to decide new selections. This type of sharing is sensible if we want to extend but not overwrite a previous sprint; we should create a different copy and add more trials. The new and old sprints must share the same model and optimization configurations with compatible model initializations. 
    \item[Cold Priming:] Cold priming a new sprint entails sharing the top-N trial HP sets of an existing sprint and recalculating the scores. This is a way of initializing the acquisition model of the new study with suggested good HP sets but without relying on the prior study's score calculation. Cold priming is helpful when transitioning to higher fidelity or changing the HP space. It is also useful when relaxing or unfreezing dimension ranges. The two sprints must still share the same model with compatible initializations.   
\end{description}

Using sprint results to prune dimensions inherently risks missing better-performing regions of the hyperparameter landscape that have yet to be explored.

The following hypotheses inform our rationale for the methodology and the experiments we conduct:

\begin{enumerate}
\item 
    Top-performing trials generally improve as fidelity levels increase, but the pattern of this increase is particular to the model configuration. Consequently, ranking different model configurations by their low-fidelity scores to project higher-fidelity ranking can be unreliable. \autoref{tab:expA-results} demonstrates this misalignment of rankings of different fidelity scores shown in parentheses. Based on the low-fidelity score, one could easily misjudge which architecture is more promising. 
\item
    Contrary to scores, optimal hyperparameter distributions are more consistent across fidelity levels. Also, fidelity reduction achieved through rotating subsets rather than by reducing epochs introduces a form of cross-validation whereby hyperparameter variances indicate how consistently training-validation subset pairs perform. This lends a degree of robustness and a measure for detection. To confidently interpret such a signal from the variance, it is pertinent to change one thing at a time, e.g., change the fidelity while keeping the dimensions consistent or the other way around.  
\item 
    Visually and analytically, the score-value distribution for each hyperparameter provides a variety of clues. Well-distributed values over the allowable range (high entropy) suggest exploratory efficiency. The concentration of top-performing values suggests the degree of confidence in convergence. Concentration around multiple optima may indicate highly correlated hyperparameters that may decrease the effectiveness of the optimization. For example, if task weights and learning rates are optimized in a multi-task model, similar weight adaptation may be achieved through a low weight and a high lr-rate or a high weight and low lr-rate, leading to unproductive HPO iterations. Examining pair correlations and domain insight in these situations may help remove redundant HPs or impose constraints, e.g., the sum of task weights must be 1. Finally, optimal value concentration at either range boundary may indicate that optimal values lie outside the allowed range.     
\end{enumerate}

\begin{table*}
\centering
\begin{tabular}{llrlcrlcrl}
        \multicolumn{8}{c}{\textbf{Experiment A: Configuration Results}\footnotemark[1]} \\[0.5ex]
\toprule
\textbf{Model} & \textbf{Grouping} & \multicolumn{2}{c}{\textbf{1E}} && \multicolumn{2}{c}{\textbf{Warm}} && \multicolumn{2}{c}{\textbf{25E+Cal}} \\
\midrule
Long-DTL\footnotemark[2] & LR0-L2 & 22.21 & (6) &  & 36.91 & (1) &  & 49.64 & (1) \\
Long-DTL\footnotemark[2] & GLOBAL & 18.12 & (12) &  & 34.35 & (10) &  & 49.63 & (2) \\
Long\footnotemark[2] & GLOBAL & 20.64 & (9) &  & 33.77 & (11) &  & 48.65 & (3) \\
Bert-DTL-POS & LR0-L2 & 13.11 & (15) &  & 36.53 & (2) &  & 48.14 & (4) \\
Bert & GLOBAL & 23.11 & (4) &  & 31.92 & (15) &  & 48.00 & (5) \\
Bert-ASL & LR0-L2 & 19.13 & (10) &  & 34.95 & (7) &  & 47.91 & (6) \\
Bert-POS & LR0-L2 & 24.19 & (2) &  & 35.47 & (6) &  & 47.84 & (7) \\
Bert-ASL-POS & GLOBAL & 27.15 & (1) &  & 35.80 & (4) &  & 47.79 & (8) \\
Long\footnotemark[2] & LR0-L2 & 21.58 & (8) &  & 35.78 & (5) &  & 47.62 & (9) \\
Bert-DTL & LR0-L2 & 21.61 & (7) &  & 36.08 & (3) &  & 47.34 & (10) \\
Bert & LR0-L2 & 15.81 & (14) &  & 34.90 & (8) &  & 46.64 & (11) \\
Bert-DTL-POS & GLOBAL & 13.09 & (16) &  & 32.47 & (13) &  & 46.64 & (11) \\
Bert-ASL & GLOBAL & 22.90 & (5) &  & 34.51 & (9) &  & 46.50 & (13) \\
Bert-POS & GLOBAL & 23.71 & (3) &  & 33.39 & (12) &  & 46.48 & (14) \\
Bert-DTL & GLOBAL & 18.24 & (11) &  & 32.42 & (14) &  & 46.29 & (15) \\
Bert-ASL-POS & LR0-L2 & 17.41 & (13) &  & 30.85 & (16) &  & 46.13 & (16) \\
\bottomrule
\end{tabular}
        \caption{F1-micro test scores of 16 model configurations determined by encoder transformer type, enabled options (DTL, POS, ASL, or None), and parameter grouping employed. Three sprints are executed. Column \emph{1E} shows single-epoch train-validation fits, executed with gp\_minimize, employing low data fidelity subsets of one \nth{6} of the training and one \nth{3} of the validation datasets, and having the lr-scheduler disabled. The dimension space is then pruned to the top-10 result range plus a small margin. Column \emph{Warm} shows F1 micro scores of the same data fidelity as \emph{1E}, again using gp\_minimize but with the lr-scheduler on, stretching through the warmup epochs (7 or 8) of the one-cycle schedule and using the pruned space derived from the single epoch result. Columns \emph{25E+Cal} contain validation F1 scores of the full-fidelity 25-epoch training-calibration-validation cycle under an Optuna study of 9 trials with Hyperband, cold primed with the top-3 HP configurations of the Warmup sprint. The quantities in parentheses show the rank for each of the score columns. 
        }
        \label{tab:expA-results}
\end{table*}

We now describe our experiments. Experiment A (\autoref{tab:expA-results}) follows the baseline methodology, considering the two transformer architectures, i.e., Bert and Longfromer. It compares two schemes of parameter grouping, GLOBAL, with one global learning rate and one weight decay rate, and LR0-L2, with separate independent learning and weight decay rates for five parameter groups. We also consider various combinations of Asymmetric Loss (ASL), Dynamic Task Loss Weighing (DTL), and the use of Part-of-speech codes (POS) as a linguistic augmentation of the input text. 

The table shows the combinations we optimized and tested, as well as the results and ranking for the three HPO phases. 

\footnotetext[1]{Ranking of values in column shown in parentheses}
\footnotetext[2]{All Longformer models use POS codes to define global attention}

\paragraph{The experiment} has three stages: (a) A single-epoch low-fidelity sprint with the lr-scheduler disabled, intended to discover the likely HP space. It uses rotating subsets of one \nth{6} of the training and one \nth{3} of the validation datasets, with a Gaussian Process surrogate to model the objective function, employing the \emph{gp\_minimize} implementation. We ran 120 iterations (trials), the first half of which were random selections to initialize the surrogate GP model. We then pruned each dimension to the respective value ranges of the top 10 iterations. We widened this interval by a small margin, i.e., for log-uniform types, such as learning and weight decay rates, we used a factor of $\pm 1.5$; for uniform numeric, such as weights, we used a small constant $0.01$; for categorical types, we reduced the categories to those that appeared in the top-10 iterations. The HP lr-warmup was frozen since we had the lr-scheduler disabled. (b) The modified dimensional space with the same data fidelity was used in the second sprint. The lr-scheduler was enabled, and the max-epoch value was set to its proper value of 25 to avoid any schedule compression distortions. The length of the run was cut off after the end of the warm-up period, which was now one of the optimized HPs that was bounded between 6 and 8 epochs. The idea was to determine the optimal values for \emph{max\_lr}. This second sprint was also executed using \emph{gp\_minimize} for 90 iterations, with the first 30 iterations randomly selecting parameters from the restricted dimension ranges determined in the prior stage. (3) The last sprint was executed on Optuna TPE with the Hyperband pruner in serial mode since we only have one GPU. It was full-fidelity, on complete datasets for 25 epochs, and included a threshold calibration phase. The study was cold-primed with the top 3 iterations from the previous stage results, and the scores were recalculated using the full dataset. We allowed nine trials.  

    The original Hyperband Algorithm \citep{li2018hyperband} is shown in \ref{alg:hyperband}.  The values we used and the correspondence to Optuna parameters are:

\begin{description}
    \item[$R$:] Denotes the maximum resource allocated per trial, measured in epochs. We train all configurations for 25 epochs, followed by 7 epochs to calibrate thresholds. We use a custom callback for pruning, which enumerates training epochs in increments of 10 which follow the pruning decision of the algorithm, while validation epochs increase by 1 and never prune. The Optuna parameter \emph{max\_resource} is set to ``auto''. The exact value of R is 32, if one includes the 7 calibration epochs. 
    \item[$\eta$:] Downsampling rate that controls how aggressively the configurations are reduced in each round of the Hyperband process. We use the default value, 3, which implies a reduction by a third. In Optuna terminology, this parameter is called \emph{reduction\_factor}.
    \item[$s_\text{max}$:] The number of brackets ends up being $s_\text{max}=3$.
    \item[$B$:] is 128 and denotes the total budget in epochs.
\end{description}

The last two values are implicitly derived by Optuna, and we assume they are calculated using these formulas:
    \begin{align*}
        s_\text{max} &= \lfloor \log_\eta(R)\rfloor & \\
        B &= (s_\text{max} + 1) R & \text{Toral Budget} 
    \end{align*}

\begin{algorithm*}
\caption{Hyperband Algorithm \citep{li2018hyperband}}\label{alg:hyperband}
\begin{algorithmic}[1]
\State Initialize $R$, $\eta$
\State $s_{\max} = \left\lfloor \log_{\eta}(R) \right\rfloor$
\State $B = (s_{\max} + 1) \cdot R$
\For{$s \in \{s_{\max}, s_{\max}-1, \ldots, 0\}$}
    \State $n = \left\lceil \frac{B}{R} \cdot \frac{\eta^s}{(s+1)} \right\rceil$
    \State $r = R \cdot \eta^{-s}$
    \State Let $T$ be the set of $n$ configurations
    \For{$i = 0, \ldots, s$}
        \State $n_i = \left\lfloor n \cdot \eta^{-i} \right\rfloor$
        \State $r_i = r \cdot \eta^i$
        \State Evaluate $n_i$ configurations with $r_i$ resources
        \State Select the top $\left\lfloor n_i / \eta \right\rfloor$ configurations based on performance
    \EndFor
\EndFor
\end{algorithmic} 
\end{algorithm*}

\section{Result Summary} 
The task we focused on is \href{https://paperswithcode.com/task/joint-entity-and-relation-extraction}{``Joint Entity and Relation Extraction''} as shown in \emph{Papers with Code}. We use the strictest criteria, following the original JEREX method, both entities and their type must be correctly identified for a relation to be considered correct. 

Since our benchmarks use the revised  Re-DocRed dataset, all variants in JEREX-L benefited when compared with the Doc-Red benchmarks shown \href{https://paperswithcode.com/sota/joint-entity-and-relation-extraction-on-3}{here}. 

The LongFormer-based model with Dynamic Task Loss Weighting (DTL) showed the best improvement reaching scores close to $50$\% with both global and partitioned parameters. The same model with static task loss weighting and global parameters was the next best model. The substantially lower score when parameters were partitioned is curious, possibly having to do with the complexity of parameter partitioning impaired by the naive weighting mechanism.

The rest of the tests all used Bert, the same transformer as the original JEREX. The best one, uses partitioned parameters, employs dynamic loss weighting (DTL) and is aided by the part of speech (POS) augmentation. These two enhancements are the ones we saw in Long-DTL which uses part of speech to define its global attention. Although not used in identical ways, both modes employ the same two enhancements POS and DTL. 

An interesting observation is that DTL with partitioned parameters seems to boost performance while the opposite happens when used with global parameters. 


\section{Conclusion} 



\bibliography{anthology,custom}
\clearpage
\newpage
\appendix


\section{Hyperparameter Optimization (HPO)}
\subsection{Hyperparameter Selection}

\subsection{Background}

Partitioning model parameters into distinct groups with separate learning rates or other optimization hyperparameters is a familiar technique in image and natural language models. As illustrated in \autoref{tab:paramGroups}, our parameter partitioning strategy focuses on learning rates (LR). The coarsest level (LR-0) divides the model into five distinct top-level groups, four corresponding to specific tasks and one to the shared transformer component. For a more granular partitioning, we could break down the shared transformer parameters into two subgroups, as shown in the next column (LR-1): one for the embedding module and another for the encoder. The encoder parameters can be further categorized based on the transformer layers (LR-2).

Additionally, we distinguish two L2 subgroups for each LR group, one containing parameters subject to L2 regularization and the other exempt from it. In keeping with the original JEREX implementation, embeddings, biases, and layer normalization parameters are excluded from L2 regularization corresponding to the sub-groups with "No" in column L2.






We performed Bayesian optimization experiments using the following specific partitioning strategies:
\begin{description}
    
    \item[Global:] This is the method used in the original JEREX. One learning rate applies to all groups, and one weight decay rate applies to groups subject to L2 regularization. Our tuning experiment optimizes the learning rate, weight decay, and the four weights that scale the loss for each task: mention localization, coreference resolution, entity classification, and relation classification. 
    \item[LR-0-L2:] Maps five learning rates for the LR-0 level groupings shown in \autoref{tab:paramGroups} and five weight decay rates to the respective L2=Yes parameter subsets. 
\end{description}

\begin{figure}
    \centering
    \includegraphics[width=1.0\linewidth]{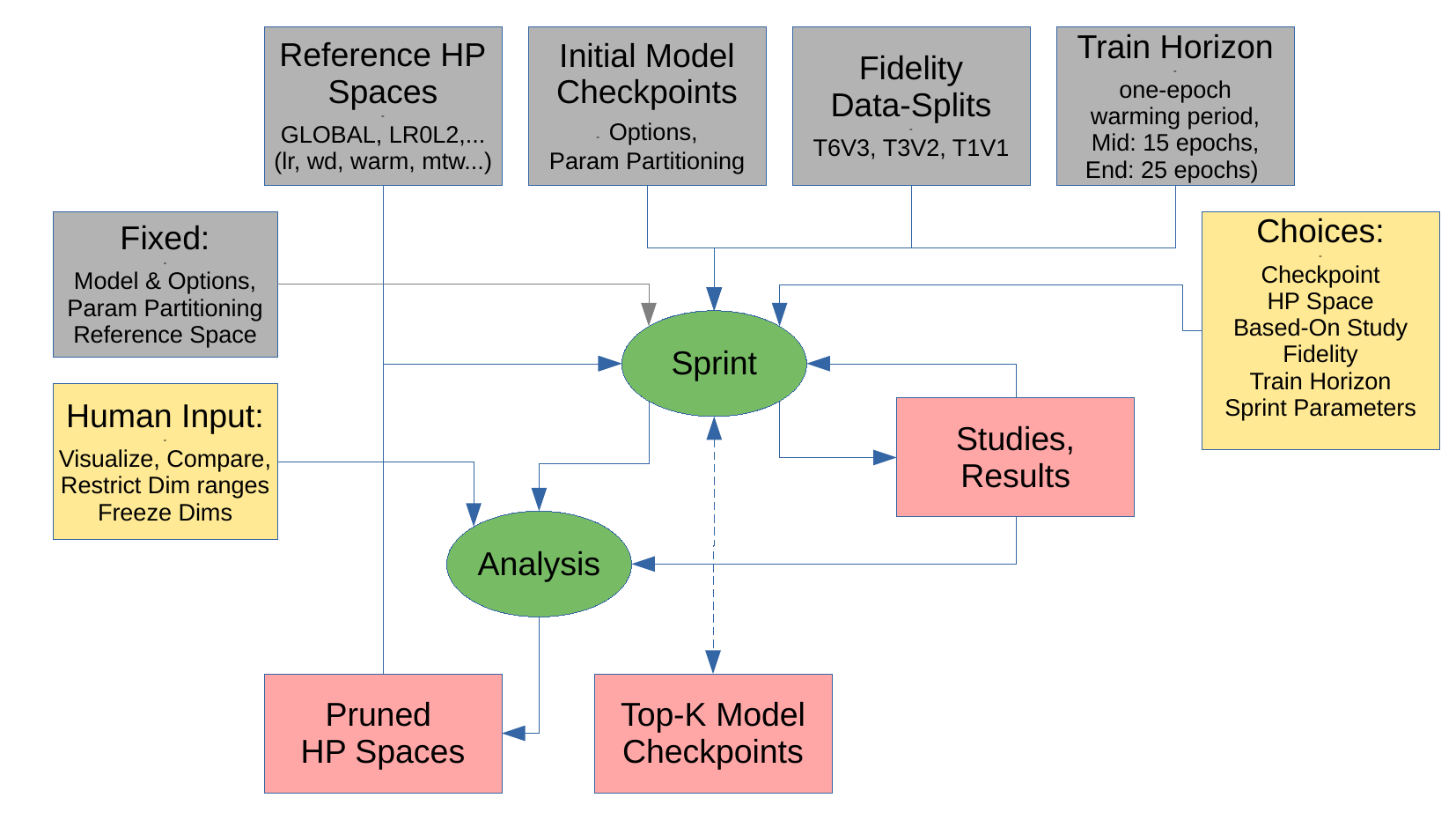}
    \caption{General framework for HPO experiments. Gray boxes denote external resources that remain fixed. Each ``Sprint" uses a model checkpoint and HP space definition to execute a given number of trials (iterations). Each trial entails fine-tuning the model at a chosen fidelity level with a proposed set of HPs suggested by a Bayesian, history-aware HPO process. The trial HPs and score results are analyzed by a human who can restrict and store pruned HP spaces for subsequent sprints to use. Sprints may optionally retain checkpoints for the top-k trials, which can be used to initialize the model in downstream sprints.}
    \label{fig:HPO-framework-1} 
\end{figure}

This paper is a case study on Hyperparameter Optimization (HPO) utilizing a framework, conceptually shown in \autoref{fig:HPO-framework-1}, and a methodology for complex natural language machine learning models using cyclic lr-schedulers and optimizer parameter partitioning. Our human-guided approach revolves around individual mini-sessions of hyperparameter exploration and refinement we refer to as \emph{sprints}\footnote{Not to be confused with Agile sprints}. Each sprint is a focused endeavor, utilizing a specific subset of data and model architecture. Following the optimization, results are analyzed, the HP space is pruned if needed, and the next sprint is planned. We use two available Bayesian optimization environments: (1) The Gaussian process based \emph{gp\_minimize} from the Scikit-Optimize (skopt) toolkit \citep{scikit-optimize}; (2) The Optuna tools, with the TPE sampler and the Hyperband pruner \citep{akiba2019optuna}. We also developed a prototype visualization application with a space-pruning aid to help progressively narrow HP ranges based on their score distribution. 

The following features characterize our approach:

\paragraph{Multi-Fidelity:} We adopt two methods of scaling the fidelity of model configurations. The first involves dividing the training dataset into $k_T$ subsets and the validation into $k_V$ subsets and rotating their use through our trials. We use the two denominators to identify data fidelity. For example, the identifier ``T6\_V4", indicates that one \nth{6}
 of the training set and one \nth{4} of the validation set is used. The second method is by restricting the number of epochs. This granular approach allows us to finely tune the model in sprints (studies) of varying data scopes
\paragraph{Trial Pruning:} For multi-epoch higher fidelity configurations we use the Optuna Hyperband pruner \citep{li2018hyperband} to stop unproductive trials early. Epoch-end validations provide the intermediate scores.
\paragraph{Hyperparameter Space:} The hyperparameters (HP) that are subject to optimization constitute a multi-dimensional HP space with specified ranges of values. Several such spaces may be preconfigured, and each experiment must be associated with one of these spaces. For example, the same model variant may have two experiments, one with a global learning rate and weight decay, while the other may optimize an array of separate learning rates and weight decays for each of several optimizer parameter groups.  

\paragraph{Sprints and Threads:} We call sprint a single HPO action involving a fixed number of trials/iterations on a specific model configuration at a given fidelity, which is manifested as a gp\_minimize or Optuna session followed by a human-guided review and preparation of a followup sprint. 
A thread is a sequence of interdependent such sprints on a fixed model configuration. Sprints in a thread can be related in several ways:
\begin{enumerate}[label=(\alph*)]
    \item Adjust HP Space: Subsequent sprints may prune their HP dimension space by freezing dimensions to confident values, or restricting dimension ranges based on prior sprint results. When the best values hover around the low or high bound, dimension ranges can be expanded. These changes are initiated by human action, allowing for a more targeted exploration in subsequent iterations. 
    \item Replace or Extend: Sprints can either overwrite a previous sprint or extend it by copying the top-k or all HP sets and their respective scores from a prior sprint. The new sprint does not recalculate the score but uses this history when selecting HP sets for new trials/iterations. This is typically a safe practice when both sprints are on the same fidelity model. This instrumentation facilitates smaller units being strung together, enabling small adjustments in between, rather than committing long periods and resources at once.
    \item Cold priming: A new sprint gets primed with the top-k HP sets from a prior sprint but with no scores carried over. This forces the new sprint to recalculate the given HP sets and derive the score. This approach is appropriate when the prior sprint uses a fidelity different from the new one. 
\end{enumerate}

 Threads should stay isolated from other threads and not be primed with HP sets or use adapted dimension spaces originating from outside their boundaries. However, different threads can be compared at corresponding stages of their evolution, focusing on the more promising ones. We consequently recommend a breadth-first strategy, whereby all threads progress through stages in tandem.

\paragraph{Initialization from Checkpoints:} Optimal effective values of hyperparameters change as training progresses. After all, this is the principle behind lr schedulers. We can more precisely tune HP schedules to follow optimal trajectories if we can understand their effective value trends through training. We discover the shape of such trajectories running HPO sprints on a given model configuration initialized from checkpoints taken at initial, mid, or late training epoch ends. 

\paragraph{Organization and Nomenclature:} We follow consistent sprint nomenclature with the following elements: (a) Model Type, e.g. \emph{JointMRC}, (b) Model variant and options, e.g. \emph{Bert-MTW}, (c) Parameter Partitioning Scheme, e.g. \emph{GLOBAL}, (d) fidelity designation e.g. \emph{T6\_V3\_M25}, (e) Initialization checkpoint epoch/step, e.g. \emph{E0\_S0} (f) customized suffix and version.  
\paragraph{Multi-phase Scheduling Strategies:} Modern optimizers and Multi-phase lr-schedulers, such as cyclic or one-cycle, modify the effective values of learning rates, momentum, or L2 weight decay with every optimizer step. These changes may become too rapid or be dramatically magnified when using low-fidelity models where the scheduled HP changes are squeezed into a fraction of the steps of the full cycle. To avoid this trap, we keep \emph{max\_epochs} to the same value used in full fine-tuning but stop the process early. To optimize the all-important max-lr parameter of the scheduler, we introduce a callback for early stopping at the end of the epoch where the warmup period ends. 

\paragraph{Single Epoch Sprints:}
Another method to avoid distortions with mult-phase schedulers, which work from any checkpoint in the fine-tuning trajectory and are immune to fidelity changes, is to run a single-epoch sprint with the lr-scheduler turned off. This yields the optimal effective learning and decay rates at the checkpoint state used to initialize the model for the sprint. This state can be the pre-trained initial state, the end of the warmup period, or a mid or late epoch, thus revealing a trend of optimal effective values for each hyperparameter through the different phases of the training cycle. We can then work backward from those optimal effective rates and calibrate our schedule to the desired shape. If the trend is noisy, we may bias our choice towards the period where the relevant HP is most important. Such meticulous refinements could lead to overfitting and should be approached with caution. 
\paragraph{Priming while Restricting Space:} When sprint results from one sprint are used in another while at the same time adjusting dimensions, we filter out the trials/iterations that don't fit the restricted new space. This ensures that the new sprint will be primed only with trials consistent with its space definition.


This structured, iterative approach to HPO, encompassing sprints, threading, and multi-fidelity exploration, offers a nuanced and effective strategy for optimizing complex machine learning models. It allows for a detailed and adaptable optimization process, enhancing model performance and efficiency.

\section{Threshold Calibration}
\subsection{Thresholding}

Our approach employs the micro-averaged relation-$F_1$ for scoring.   

Notably, training does not involve thresholds. Classifiers are independent of each other, depending only on the output of a shared encoder. The sampling process controls the input to each classifier, which contains all positive instances along with a random sample of negative ones. The entire classification distribution is passed into the computation of the respective task losses. 

In contrast, inference considers all candidate spans and span pairs that flow sequentially through the tasks. Each of the three multi-label classifiers filters its outcome using a threshold and forwards only positive predictions to the next task. 
Since our primary optimization objective is to maximize the relation $F_\beta$ score, which becomes known at the end of the task chain,  the underlying computation graph would make it too complex to express a differentiable loss function with respect to thresholds. This discourages us from pursuing an end-to-end gradient-based threshold optimization solution. It is possible, though, to use a deterministic optimization method for the first encountered threshold, the mention threshold, in a way that it maximizes the mention $F_\beta$ score.

We also provide a validation-based hill-climbing threshold calibration strategy that can be applied to any combination of thresholds. In contrast to our SCut optimization which maximizes the mention micro-$F_\beta$ score, our Hill-Climbing algorithm maximizes the relation-$F_\beta$, which is our primary objective, and is presented in \ref{sec:hill-climbing} below. 

\subsubsection{Mention Autho-threshold}\label{sec:mention-autothrsh}

A new threshold management module and meta-learner incorporates three optional capabilities. 
The first  calculates the optimal mention threshold analytically, during evaluations, based on the general algorithm for SCut threshold optimization \cite{Fan2007ASO}. It entails aligning candidate spans $s_i$ with predicted probability $p_i$ and an indicator $t_i$ denoting whether the span represents a true mention according to the dataset annotations as shown in equation \ref{eq:scut_gt} where $\mathcal{T}^M$ is the set of true mentions.  This list of triplets  $\left< s_i,p_i,t_i\right>$ is ordered by probability as shown in equation \ref{eq:scut_order}.  $N$ is the count of all spans and $N^T=$ is the count of those that are true mentions. Each validation step  collects true positive (TP), false negative (FN) and false positive (FP) counts. At the end of the epoch, all entries are ordered by probability, and the point maximizing the aggregate \emph{mention-F1-micro} score determines the optimal threshold. We tested alternative metrics $F_2$ through $F_4$ that favor recall, to see if allowing more mentions to flow to subsequent classifiers may be helpful, but $F_1$ proved to be the best choice. To reduce the cost of the algorithm we discard all instances with a probability lower than the lowest true mention since none of these entries contributes to the true positive sum (the numerator) and thus doesn't affect the computation. 
\begin{align}
    t_m &=  \begin{cases}
        1 & \text{if } s_m \in \mathcal{T}^M \\
        0  & \text{otherwise}
    \end{cases} \label{eq:scut_gt} \\
    p_m < p_n &\implies m < n \label{eq:scut_order}\\
    N^T &= \sum_{\forall i} t_i & \text{True mentions} \nonumber \\
    N &= \sum_{\forall i} 1 & \text{all predictions} \nonumber \\
    \text{FN}_m &=  \sum_{i < m} t_i  \label{eq:scut_FN}\\
    \text{TP}_m &=  N^T - \text{FN}_m  \label{eq:scut_TP}\\
    \text{FP}_m &=  N - m - \text{TP}_m \label{eq:scut_FP}
\end{align}

Equations \ref{eq:scut_FN} to \ref{eq:scut_FP} return metrics for a given point $m$. Since FN$_m$ depends on range $i < m$, the vector $\overline{\text{FN}}$ is a cumulative sum of vector $\bar{t}= [t_1,t_2,\dots,t_N]$ as shown in \ref{eq:scut_vecFN}. Equations \ref{eq:scut_vecTP} and \ref{eq:scut_vecFP} express 
TP and FP in vector notation.

\begin{align}
    \overline{\text{FN}} &= \text{cumsum}(\overline{t}) \label{eq:scut_vecFN}\\
    \overline{\text{TP}} &= N^T - \overline{\text{FN}} \label{eq:scut_vecTP}\\
    \overline{\text{FP}} &= N - \text{cumsum}(\overline{\mathbf{1}}_N) -      
        \overline{\text{TP}} \label{eq:scut_vecFP} \\
    \overline{F_\beta} &=  \frac{(1+\beta^2) \overline{\text{TP}}}{
    (1+\beta^2) \overline{\text{TP}} + \beta^2 \overline{\text{FN}} + \overline{\text{FP}}} \label{eq:scut_vecF}\\
     \zeta^{*M} &= p_{[i^*]} \mid i^* = \underset{i}{\arg\max} \overline{F_\beta}[i] \label{eq:scut_best_thresh}
\end{align}

$\overline{\mathbf{1}}_N$ represents a unit vector of length $N$, whose cumulative sum is the vector of sequential numbers from 1 to $N$. Equation \ref{eq:scut_vecF} computes the vector of all F$_\beta$ scores. The optimal mention threshold $\zeta^{*M}$ is the probability that corresponds to the point with the highest score shown in equation \ref{eq:scut_best_thresh}. At the end of the epoch, we update the model's mention threshold. 

We found that this approach works well in epoch-end evaluations during training because it is accurate and efficient. For threshold calibration after training, we favor an iterative hill-climbing, alas slower, method that optimizes thresholds with respect to the relation $F_1$-micro score, our primary objective. 

It is worth noting that although mapping candidate mentions to the corresponding ground truth is straightforward, such mapping for coreferences, entities and relations  is complex, and costly. This is the reason that we use analytical optimization only for the mention threshold.

\subsubsection{Hill-climbing Meta-learner}\label{sec:hill-climbing}
The second capability of the Threshold Management Module optimizes any or all of the thresholds and was introduced to enhance the original model's fixed threshold approach. We implemented a calibration phase that adjusts the thresholds based on the model's current state with respect to the validation dataset. These calibrations may occur at the end of each training epoch to optimize thresholds before model selection evaluations. They can also be conducted after the end of training to establish the best thresholds to use for inference.

During the calibration process, a meta-learner iterates over epochs of validation data, employing a hill-climbing procedure considering multiple values per threshold to find the best combination that maximizes the \emph{rel-nec-F1-micro} score\footnote{This score considers both the correctness of the relation and the accurate prediction of the classes of the two related entities}. Flexibility in choosing which thresholds participate in this optimization for each calibration epoch allows us to structure a variety of strategies. 

A rotational strategy is less costly and suitable for epoch-end evaluations during training. This strategy cycles through the three thresholds in consecutive epochs, optimizing one at a time. The model's forward evaluation method has been modified to accommodate a list of one\footnote{for thresholds that are not being optimized, retaining the current value} or three values per threshold, and to efficiently process an ensemble of all value combinations. Each combination is scored at the end of the epoch when the highest score determines the optimal values for the respective thresholds. More accurate strategies that optimize two or all three thresholds per calibration epoch are possible but computationally more costly as they have to compute 9 or 27 scores per validation step, respectively. 

Let's consider the most complex scenario, where all three thresholds (mention, coreference, and relation) are optimized concurrently during the same calibration epoch. The basic idea is to try three values for each threshold and calculate scores for all 27 combinations. The three values are chosen so that the first is always the current threshold value $\zeta_0$, while the other two are low and high perturbations\footnote{The value of $delta$ is configurable} $\zeta_0 \pm\delta$. In our experiments, we configure $\delta=0.05$ during training evaluations and $0.025$ in post-training calibrations. The new threshold values are determined by selecting the combination that maximizes the score using the following equations:

\begin{align}
 S_{i,j,k} &= F_1(\zeta^M_i,\zeta^C_j,\zeta^R_k) \mid \forall i,j,k \label{eq:thresh_S}\\
    i^*,j^*,k^* &= \underset{i,j,k}{\arg\max} \, S_{i,j,k} \label{eq:thresh_ijk} \\
 {\zeta'}^M_0 &= \zeta^M_{i^*} \label{eq:thresh_M}\\
{\zeta'}^C_0 &= \zeta^C_{j^*} \label{eq:thresh_C}\\
{\zeta'}^R_0 &= \zeta^R_{k^*} \label{eq:thresh_R}.
\end{align}
In equation \ref{eq:thresh_S}, the $F_1$ function is applied to all combinations of values, resulting in 27 scores indexed by the three threshold indices. Equation \ref{eq:thresh_ijk} identifies the indices $i^*,j^*,k^*$ corresponding to the three threshold values resulting in the highest score. Finally, equations \ref{eq:thresh_M} to \ref{eq:thresh_R}  update each threshold with its new value $\zeta'$. Each calibration epoch progressively yields thresholds that improve the \emph{relation-F1-micro} score until no further improvement is observed or a predefined iteration limit is reached.

\subsubsection{Class-specific Relation Thresholds}

Equation \ref{eq:thresh_R} assumes the relation threshold is global, i.e., one threshold value applies for all relation classes. We introduced the option of replacing the original global threshold with a vector of relation thresholds independently optimized for each relation class. This requires that we calculate class specific $F_1$ scores. Although each different class $r$ could theoretically be optimized using different values for $\zeta^{M}$ and $\zeta^{C}$, we choose to keep these optimized based on the total F1 micro score while allowing the relation threshold $\zeta^{R}$ to vary based on the relation-class specific $F_1$ scores. We refine equations \ref{eq:thresh_S} through \ref{eq:thresh_R} accordingly: 

\begin{align}
 S_{i,j,\bar{k}} &= F_1(\zeta^M_i,\zeta^C_j,\overline{\zeta^R_{\bar{k}}} ) \mid \forall i,j,\bar{k} \label{eq:thresh_S2}\\
\overline{S_{i,j,\bar{k}}} &= \overline{F_1}(\zeta^M_i,\zeta^C_j,\overline{\zeta^R}_{\bar{k}}) \mid \forall i,j,\bar{k} \label{eq:thresh_Sr}\\
     i^*,j^*,k_q^* &= \underset{i,j,k_q}{\arg\max} \, S^q_{i,j,k} \label{eq:thresh_ijkr} \\
 {\zeta'}^M_0 &= \zeta^M_{i^*} \label{eq:thresh_Mr}\\
{\zeta'}^C_0 &= \zeta^C_{j^*} \label{eq:thresh_Cr}\\
{\zeta'}^{Rk}_0 &= \zeta^R_{k^*} \label{eq:thresh_Rr}.
\end{align}

\subsubsection{Implementation Notes}

We built an auto-threshold module for the joint multi-instance entity-relation extraction model with the following functionality:
\begin{itemize}
    \item Holds and maintains the current threshold values, which are initialized from the respective command arguments and updated according to the thresholding strategy followed. Whenever the model requires threshold values, it gets them from the ThresholdManager class instance. We adapted the original JEREX code to accommodate all capabilities described here for models of class \emph{JointMultiInstanceModel}.
    \item Auto-thresholding strategies enable automatic threshold tuning, independently for each threshold, through parameters that control epoch-start/stop/skip and a perturbation $\delta$.
    \item Before an epoch-end evaluation starts, qualifying thresholds receive a list of three values instead of one.  Concretely, given the last epoch's threshold value $\zeta$, the new value $\zeta'$ is the one yielding the best relation-F1 score from comparing the last value $\zeta$ and its two perturbations of magnitude $\delta$:  \[ \zeta' = \underset{x}{\arg\max} \, F_1(\hat{y}_z,y) \mid z \in \{\zeta,\zeta \pm \delta\} \] evaluated at the end of an epoch.  
    \item The evaluation results are susceptible to fluctuations of the Mention Threshold, so we provide the option to auto-compute optimal values based on the Mention Localization F1, F2, F3, or F4 scores. The auto-compute option is designed for the early epochs to determine a good starting value, and it is ignored after the first epoch when auto-thresholding is activated.
\end{itemize}

Our implementation of the meta-learner leverages the original JEREX mechanism of accumulating step metrics, storing them on the file system, and loading them back at the end of the epoch for scoring. As mentioned earlier, the hill-climbing meta-learner entails two modifications of the original model. The first involves splitting the \emph{forward} method of the joint model into four sections, i.e., transformations independent of any threshold, those that are dependent on the mention threshold, those dependent on mention and coreference thresholds, and those dependent on all thresholds. Instead of a single value, each threshold accepts lists of values, and the algorithm avoids unnecessary recalculations. For example, transformations involving a mention threshold are executed once, sharing results with each of the coreference-dependent transformations, also performed once for all relation thresholds. 
The second addresses the repeated execution of the relation calculations that can become extremely slow. To accelerate the validation, the function \emph{create\_local\_entity\_pairs} was rewritten, retaining the content and order in the resulting tensors but reducing its cost to a quarter or fifth of the original implementation.

\subsection{Bayesian Hyperparameter Tuning}
\label{sec:autotuning}

\subsection{Fine-Tuning Process} \label{sec:autothreshLoop}

All fine-tuning runs follow a consistent process shown in Algorithm \ref{alg:FineTuning}.
The notation we use is shown in \autoref{tab:ftNotation}.

\begin{table}
    \centering
    \begin{tabular}{r|l}
     $E$    &  Total Number of Epochs\\
     $C$    &  Max Calibration iterations \\
     $\mathcal{D}^{tr}$ & Training dataset \\
     $\mathcal{D}^{val}$& Validation dataset \\
     $\mathcal{D}^{test}$& Test dataset \\
     $\theta$ & Model parameters (state) \\
     $\theta^*_i$ & Best performing model as of i \\
     $\zeta^{m}$ & Mention threshold \\
     $\zeta^{c}$ & Coreference threshold \\
     $\zeta^{r}$ & Relation Threshold \\
     $\delta$ & Perturbation magnitude\\
     $Z_i^{m}$ & candidate set at i $\{\zeta_{i-1}^{m}, \zeta_{i-1}^{m}\pm\delta\}$ \\
     $Z_i^{c}$ & candidate set at i $\{\zeta^{c}_{i-1}, \zeta^{c}_{i-1}\pm\delta\}$  \\
     $Z_i^{r}$ & candidate set at i $\{\zeta^{r}_{i-1}, \zeta^{r}_{i-1}\pm\delta\}$ \\
     $\mathcal{P}_i$ & Permutations at i 
            $ \{ (\zeta^{m},\zeta^{c},\zeta^{r})\}$ \\
            & where $\zeta^{m} \in Z_i^{m}, \zeta^{c} \in Z_i^{c}, \zeta^{r} \in Z_i^{r}$ \\
     $\mathcal{M}|\theta$ & Model at State $\theta$ \\
     $\fit$ & Train (train.fit) model $\model$ \\
     $p^{m}$ & Mention probability distribution \\
     $p^{c}$ & Coreference  probability distribution \\
     $p^{r}$ & Relation probability distribution \\
     $\val$ & Validate (train.valid) model $\model$\\
     $\test$ & Test (train.test) model $\model$ \\  
     $F^{m}_\beta$ & Mention micro-$F_\beta$ score \\
     $F^{r}_\beta$ & Relation micro-$F_\beta$ score \\
     $F^{r*}_{\beta i}$ & Best score as of the end of epoch i\\
     $\pi$ & Threshold Optimization Policy
    \end{tabular}
    \caption{Notation Used in the Fine-Tuning Algorithm}
    \label{tab:ftNotation}
\end{table}


In our standardized model training and evaluation process, we've implemented a systematic approach that encompasses various phases: training, threshold calibration, validation, and testing. This process commences with an invocation of the 'fit' method, followed by a hill-climbing calibration phase. During this phase, we conduct multiple evaluations while adjusting threshold values to maximize the model's performance.

The 'fit' function operates iteratively across $E$ epochs. In each epoch, it trains the model on the training dataset and subsequently evaluates its performance on the validation dataset. To enhance threshold calibration within the validation process, we utilize two distinct methods. One is deterministic mention threshold optimization through the SCut method, while the other employs our general hill-climbing calibration algorithm to optimize the coreference and relation thresholds.

To facilitate the hill-climbing algorithm, we've introduced a modification in the validation process. Rather than accepting a single value for each threshold, it can now optionally accept three values. This adjustment allows us to compute a performance score for every possible combination of threshold values, from which we select the best-performing combination.

The optimization strategy for thresholds is defined by a policy $\pi$, which governs various details of the process. When only one threshold receives a triplet of values, the process generates three distinct performance scores. With two threshold triplets, we exhaustively explore all combinations, resulting in nine scores. With three threshold triplets, the total number of combinations reaches 27. These threshold triplets are structured as follows: the first value remains consistent with the existing threshold value, while the subsequent two values represent perturbations achieved by adding and subtracting a small value $\delta$. For the 'fit' call, we employ $\delta=0.05$, and for the calibration phase, we use $\delta=0.025$.

Consequently, threshold calibration is applied in two distinct phases. During the 'fit' method, we perform deterministic optimization for mention thresholds, while employing hill-climbing for the other two types. To manage computational costs, we initiate the application of the hill-climbing technique after the \nth{10} epoch and limit the number of evaluated permutations within each epoch by rotating which thresholds are subject to optimization.

The ``Calibration Phase" that follows relies on the hill-climbing method. This is because our primary performance score is the relation micro-$F_1$, while the S-Cut deterministic method optimizes mention thresholds only according to the mention micro-$F_1$.
This is a relatively expensive computational cost, but it selects the best value for each threshold according to our primary score and using the validation set. These threshold values are then used to run the final test on the test dataset.

Our experiments follow this pipeline, which is repeated three times with the same hyperparameters. We report the mean of the resulting three test scores.

\begin{algorithm*}[hbt!]
\caption{Fine-Tuning with Threshold Calibrations}\label{alg:FineTuning} \begin{algorithmic}
\State{Setup and Initialize $\model|\theta_0$ \Comment{Load pre-trained or from checkpoint}}
\State Initialize $\zeta^m_0,\zeta^c_0,\zeta^r_0 $\Comment{From hyperparameters}
\State Load $\mathcal{D}^{tr}$, $\mathcal{D}^{val}$,$\mathcal{D}^{test}$ 
\State \textbf{Fit Model:} \Comment{Train-Calibrate-Validate for E epochs}
\For{$i \in [1,E]$}
    \State Train:
    \State{ $\theta_i,p^{m},p^{c},p^{r} \leftarrow \fit(\theta_{i-1},\mathcal{D}^{tr})$  \Comment{Train and retain probability distributions}}\\
    \State Prepare Thresholds: \Comment{SCut + Hill-Climbing}
    \State $Z_i^m \leftarrow \{\zeta_i^{m*}\} \leftarrow \{\text{S-Cut}[ F^{m}_1](p^{m}_i)\}$ \Comment{Optimal mention threshold maximizing $F^m_1$}
    \State{$Z^{c}_i \leftarrow \{\zeta^{c}_{i-1}, \zeta^{c}_{i-1}\pm\delta\}$ \Comment{Coref threshold trial values per $\pi(i)$}}
    \State{$Z^{r}_i \leftarrow \{\zeta^{r}_{i-1}, \zeta^{r}_{i-1}\pm\delta\}$ \Comment{Relation threshold trial values per $\pi(i)$}}
    \State{ $\mathcal{P}_i \leftarrow \{(\zeta^{m}, \zeta^{c}, \zeta^{r}) \, | \, \zeta^{q} \in Z^{q} \, \forall q \in [m,c,r] \}$
    \Comment{All permutations of trial values}}\\
    \State Validate \Comment{Single pass, evaluate for all threshold permutations}
    \State  $\{(F_1^r,\zeta^{m},\zeta^{c},\zeta^{r})\} \leftarrow \val(\mathcal{D}^{val},\mathcal{P}_i)$
    \Comment{Validation result set by threshold permutation}
    \State $F^{r*}_1,(\zeta^{m*}, \zeta^{c*}, \zeta^{r*}) \leftarrow \underset{(\zeta^{m}, \zeta^{c}, \zeta^{r})}{\arg\max} F^r_1$ \Comment{Choose Best-score permutation}
    \If{$F^{r*}_1 >$ best-score}
        \State Replace Checkpoint
    \EndIf
\EndFor
\State
\State \textbf{Calibrate Thresholds} \Comment{Iterative Hill-Climbing}
\State $\zeta^m_0,\zeta^c_0,\zeta^r_0 \leftarrow \zeta^{m*}, \zeta^{c*}, \zeta^{r*}$
\For{ j in [1,C]} 
    \State{ $\mathcal{P}_J \leftarrow \pi(j,\zeta^m_{j-1},\zeta^c_{j-1},\zeta^r_{j-1})$
    \Comment{Trial threshold permutations per Policy $\pi$}}
    \State  $[F_1^r,\zeta^{m},\zeta^{c},\zeta^{r}] \leftarrow \val(\mathcal{D}^{val},\mathcal{P}_i)$
    \Comment{Validation results for each threshold permutation}
    \State $F^{r*}_1,(\zeta^{m*}, \zeta^{c*}, \zeta^{r*}) \leftarrow \underset{(\zeta^{m}, \zeta^{c}, \zeta^{r})}{\arg\max} F_1$ \Comment{Choose Best-score permutation}
    \State $\zeta^{m}_j, \zeta^{c}_j, \zeta^{r}_j \leftarrow \zeta^{m*}, \zeta^{c*}, \zeta^{r*}$ \Comment{Set best Thresholds for the next iteration}
    \State Break if no improvement or continue
\EndFor\\
    \State \textbf{Test}
    \State Scores $\leftarrow\test(\mathcal{D}^{test})$\Comment{Test using latest Thresholds}
    \State Report Scores
\end{algorithmic}
\end{algorithm*}

\end{document}